\documentclass{article}

\usepackage[final]{neurips_2022}
\usepackage{multirow}
\usepackage{hhline}
\usepackage{adjustbox}
\usepackage{booktabs}

\usepackage[utf8]{inputenc} 
\usepackage[T1]{fontenc}    
\usepackage{hyperref}       
\usepackage{url}            
\usepackage{booktabs}       
\usepackage{amsfonts}       
\usepackage{nicefrac}       
\usepackage{microtype}      
\usepackage{xcolor}         
\usepackage{svg}            
\usepackage{amssymb}
\usepackage{amsmath}

\title{CASPR: Customer Activity Sequence-based Prediction and Representation}

\author{%
  Pin-Jung Chen\thanks{Joint first authorship} \\
  Microsoft Corporation \\
  \texttt{piche@microsoft.com} \\
  \And
  Sahil Bhatnagar\footnotemark[1] \\
  Microsoft Corporation \\
  \texttt{sabhatn@microsoft.com}
  \And
  Sagar Goyal\:\thanks{Work performed while at Microsoft Corporation.} \\
  \texttt{goyalsagar@outlook.com} \\
  \And
  Damian Konrad Kowalczyk\thanks{Joint senior authorship} \\
  Microsoft Corporation \\
  \texttt{damian.kowalczyk@microsoft.com} \\
  \And
  Mayank Shrivastava\footnotemark[3] \\
  Microsoft Corporation \\
  \texttt{mayank.shrivastava@microsoft.com} \\
}

\begin{document}
\maketitle
\begin{abstract}
Tasks critical to enterprise profitability, such as customer churn prediction, fraudulent account detection or customer lifetime value estimation, are often tackled by models trained on features engineered from customer data in tabular format. Application-specific feature engineering adds development, operationalization and maintenance costs over time. Recent advances in representation learning present an opportunity to simplify and generalize feature engineering across applications. When applying these advancements to tabular data researchers deal with data heterogeneity, variations in customer engagement history or the sheer volume of enterprise datasets. In this paper, we propose a novel approach to encode tabular data containing customer transactions, purchase history and other interactions into a generic representation of a customer's association with the business. We then evaluate these embeddings as features to train multiple models spanning a variety of applications. CASPR, Customer Activity Sequence-based Prediction and Representation, applies Transformer architecture to encode activity sequences to improve model performance and avoid bespoke feature engineering across applications. Our experiments at scale validate CASPR for both small \& large enterprise applications.
\end{abstract}
\section{Introduction}
%
Enterprises today store and process an increasingly large amount of customer data, commonly structured in databases. Utilizing big data and machine learning tools to enhance the business profitability and sustainability is an active area of interest, research and development for businesses regardless of their scale (\cite{santoro2018big}). Tabular data collected by businesses include profile information about their customers and records of various interactions with the business and products. These could include tables containing transaction information, physical visit information, online activity, customer feedback etc. Building models that capture complex customer interactions and sequences of activities allows businesses to understand their customers better; accurately predicting customers who might churn leads to better marketing and mitigation strategies, understanding products and services a customer is more likely to use via a strong product recommendation engine leads to greater engagement and mapping potential customer value ensures the businesses focus on their most profitable ventures. \indent Traditional statistical tools for analyzing tabular customer data are insufficient in modelling the complexity and interactions of data collected and rely on assumptions (\citet{Schmittlein1987, Fader2005}) of data distribution and characteristics to make predictions. Intricately engineered features to help predict specific problems such as customer churn are difficult to transfer to other problems such as recommending products and require a new feature representation for the same. The challenge of modelling customer data and the importance of predicting problems critical to businesses motivate the development of the Customer Activity Sequence-based Prediction
and Representation (CASPR) framework to enhance learning from tabular data on customer behaviour and advance real-world predictive tasks in business.
%
\section{Approach \& Technical Challenges}
%
Our proposed CASPR framework models raw tabular data of customer activities as a timestamped sequence of events, where each event represents a row of information in a table corresponding to some customer interaction. Some examples of such tables could be web activity logs, transactions logs, customer service interactions etc. The goal of the CASPR framework is to generate a latent vector representation of each customer in the database, which encapsulates all the information about the customer's interactions with the business. The transformation of raw tabular data into a journey of customer activity forms the basis of generating these vector representations.
We leverage progress made in the world of large language models using the transformer network architecture (\cite{vaswani2017attention, bert2018}) and modify the semantics of how transformer models are generally trained; events in the customer journey represent words in a sentence, where each customer journey is like a sentence. Just as language models can understand the semantics of words in a sentence, the CASPR transformer understands the semantics of events w.r.t. their position in a customer's journey.
A significant advantage of the CASPR framework is that training the model to generate latent vector representations of customers is a self-supervised process, similar to the training of large BERT models for natural language (\cite{bert2018}). The pre-training of CASPR transformer models forms the basis of their applicability across a variety of business tasks. CASPR vector representations or embeddings can be used to supplement existing models, such as collaborative filtering models for a product recommendation, or fine-tuned to predict tasks such as customer churn or lifetime value. Using CASPR prevents the need of generating complex engineered features which capture the semantics of tabular data specific to a task to solve, and also allows an increase in the scale of input raw tabular data to make more accurate predictions.
\section{Contributions \& Organization}
In this study, we propose CASPR, a novel framework for computing deep embedding representations
\begin{itemize}
    \item 
    We introduce the CASPR framework to learn latent vector representations of customer journeys extracted from tabular datasets commonly found in business scenarios. We show how this form of representation learning can reduce or eliminate the need for complex feature engineering and scales across tasks (see Section \ref{section:applications}).
    \item 
    We demonstrate the performance of a pre-trained CASPR model in helping solve multiple business applications and improve upon baseline models (see Table \ref{table:classificationchurn}).
    \item 
    We evaluate popular frameworks for distributed deep learning, to identify the most cost effective way of computing CASPR embeddings from tabular big data in an industrial setting (see Table \ref{table:distribution}).
\end{itemize}
The paper proceeds as follows: in Section 4 we review background concepts such as approaches to business problems, current tabular representation learning techniques and techniques of operationalization of deep learning models at scale in the industry. In Section 5 we detail the CASPR framework and give an overview of the business applications that may benefit from using CASPR. Section 6 describes our experiments conducted to evaluate the generalization and scalability of CASPR models in real-world scenarios. Finally, we offer our conclusions in Section 7.
\section{Background}
\subsection{ML Approaches for Business Applications}
Real-world business problems such as Customer Churn, Customer Lifetime Value (CLV), or product recommendation are critical to enterprises due to their direct impact on revenue. Before large models and high-performance computing resources became available, simple parametric statistical models were adopted to describe customer behaviour. \cite{Schmittlein1987} proposed the Pareto/NBD framework which uses Pareto and NBD distributions for modelling customer churn and representing customer purchase frequency. \cite{Fader2005} expanded on this by incorporating the popular RFM (Recency, frequency, monetary value) paradigm into the Pareto/NBD framework. With the advent of big data and machine learning (ML) algorithms, researchers started to build predictive models for these business problems. \cite{Ahmad2019} and \cite{Vanderveld2016} developed tree-based models with feature engineering for churn and CLV predictions. While these ML models had shown promising results, the need to handcraft a large number of features can become a bottleneck for maintainability and scalability. Inspired by neural embeddings, a technique pioneered in the field of Natural Language Processing (NLP), researchers began to leverage embeddings to automatically extract rich patterns of customer behaviour from raw data. \cite{Barkan2016} used skip-gram with Negative Sampling (SGNS) (\cite{Le2014}), a popular neural embedding algorithm, to develop a method called item2vec for item-based collaborative filtering for Product Recommendation. \cite{Grbovic2016} proposed several approaches that learn product representations by applying SGNS to user purchase history extracted from e-mail receipt logs. \cite{Chamberlain2017} introduced a hybrid CLV system for e-commerce that combines handcrafted features and customer embeddings learnt from browsing sessions.
\subsection{Table Representation Learning}
In real-world AI applications, the most common data type is tabular data (\cite{Chui2018, Shwartz-Ziv2021, Borisov2021}). Despite the recent success of deep learning on image, text, and speech data, tabular data remain a challenge for deep learning. Traditional ML methods such as tree-based ensemble models still dominate the tabular data domain (\cite{Borisov2021, Shwartz-Ziv2021}). Recently, several new neural architectures have been proposed, attempting to achieve performance improvement and learn meaningful representation on tabular data. \cite{Arik2019} proposed TabNet, which uses sequential attention to perform tree-like feature selection and reasoning, and demonstrated that self-supervised pre-training could significantly improve performance and led to faster model convergence. \cite{Huang2020} extended Transformer (\cite{vaswani2017attention}) and introduced TabTransformer which learns contextual embeddings of categorical features. TabTransformer matched the performance of tree-based ensemble models (GBDT) while being robust against noisy and missing data. SAINT (Self-Attention and Intersample Attention Transformer) proposed by \cite{Somepalli2021} uses a hybrid attention mechanism that performs attention over both rows and columns. Unlike TabTransformer, it projects both categorical and continuous features into a common latent space. The authors also leveraged self-supervised contrastive pre-training in the semi-supervised setting.
\subsection{Operationalization \& Distributed Processing}
In the digital world, data are generated from various sources and companies have been putting effort into collecting more and more data to better understand their customers. As the data grows and accumulates, it becomes non-trivial to process these large datasets and deploy models in production at scale without leveraging distributed systems and frameworks. Apache Spark (\cite{Zaharia2016}) is an open-source unified engine for large-scale distributed data processing that can capture streaming, SQL, machine learning, and graph processing workloads. Spark extends MapReduce with Resilient Distributed Datasets (RDDs), a distributed memory abstraction that allows users to perform in-memory computation on large clusters with performance, reliability and privacy implications (\cite{Zaharia2012,kowalczyk2018scalable}).
Horovod introduced by \cite{Sergeev2018}, is a distributed deep learning training framework that is compatible with most of the common ML libraries. It uses a custom implementation of the ring-allreduce algorithm for efficient inter-GPU communication, enabling users to easily scale a single-GPU training script to train across multiple GPUs in parallel without much code modification. Petastorm (\cite{Gruener2018}) is an open-source library which enables training deep learning models directly from Apache Parquet datasets. PyTorch Distributed Data-Parallel (DDP) (\cite{Li2020}) is a single-program multiple-data training paradigm for distributed training applications which can run across multiple machines. It replicates the model on every process and feeds each model replica with a different subset of the training samples to generate gradients independently. A gradient communication layer is applied at the end of every iteration to ensure that each model replica is consistent and synchronized.
\section{Framework Overview}
Our approach for creating a representation of raw tabular data focuses on creating encoded embedding representations of the entities in the dataset. In this section, we introduce the CASPR framework which models raw tabular data as a sequence of interaction events or activities performed by an entity and then learns representations for each entity. CASPR first transforms the tabular data into a set of sequential activities per user or entity and is then trained with objectives such as masked entity recovery. This helps the model understand the structure and semantic relationship between activities in a sequence. The generated embedding representations for each entity can then be used for a range of downstream tasks such as predicting churn, lifetime value or detecting fraudulent accounts. We perform the pretraining step for each dataset once and then reuse the generated embedding representations across tasks.
\subsection{Data preparation and pretraining}
\begin{figure}
    \centering
    \includegraphics[width=1\columnwidth]{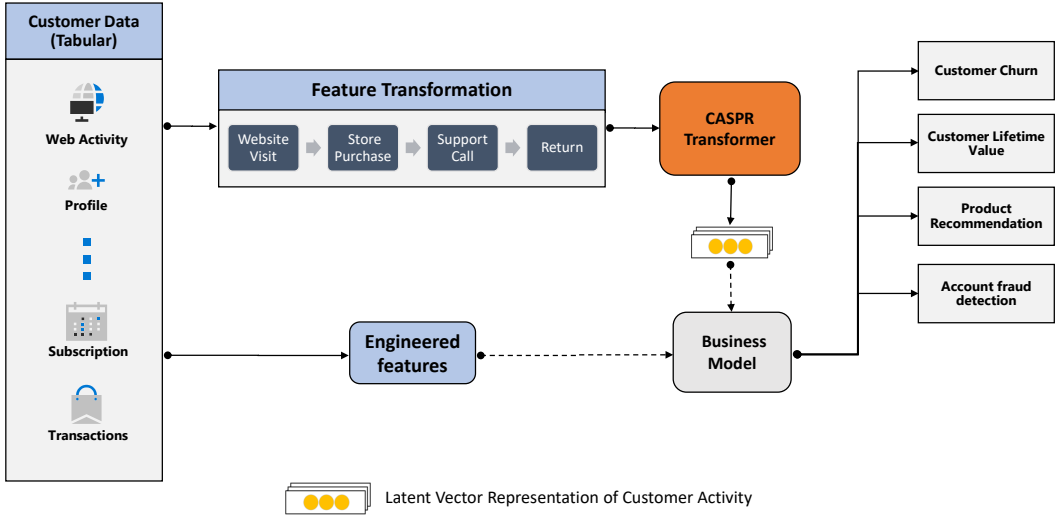}
    \caption{Overall CASPR architecture diagram. Raw rows in tabular customer data are converted to a sequence of events which are then fed into a CASPR transformer model to generate vector representations for each customer.
    Tabular features are recency, frequency and monetary features extracted from the raw customer data.
    The business model can utilize either the RFM features, CASPR embeddings or a combination of both.}
    \label{fig:e2e}
\end{figure}
In a tabular dataset $D$, we denote each row $A$ as a combination of an entity $E$, a timestamp $T$ and a set of $N$ attributes $A_E^T = \{E,T,A_1,A_2...,A_N\}$. An activity is the set of attributes $\{A_1,A_2...,A_N\}$ corresponding to a row occurring at time $T$.
The goal of CASPR is to generate a latent vector representation $\nu_E$ of all the activities associated with an entity $E$. The first step in the CASPR pipeline is to transform the raw input dataset $D$ into a sequential representation of activities for each entity $E$ ordered by the timestamp $T$. Thus, for each $E$, we generate a sub-dataset $D_E$ containing an ordered list of activity rows
\begin{math}
D_E = (A_E^1,A_E^2...,A_E^t).
\end{math}
Note that the timestamps do not need to be synchronized across different entities and activities. As a parameter of the model, we fix the maximum number of activities per entity to $t$.
%
We then batch the sub-datasets $D_E$ and pass them through a learn-able categorical embedding layer to convert all categorical activity attributes into a vector representation of dimension $d=\sqrt{cardinality(C)}$ where $C$ is a categorical attribute. Categorical attributes include strings, ordinal values and any other non-numeric data type. This net sequence of activity features is then passed into a multi-head transformer model, where the model is trained in an unsupervised manner to optimize the masked entity recovery objective. The generated entity embeddings $\nu_E$ can then be used as input features for training downstream models for different tasks.
\subsection{Model Overview}
%
\begin{figure}
    \centering
    \includegraphics[width=0.9\columnwidth]{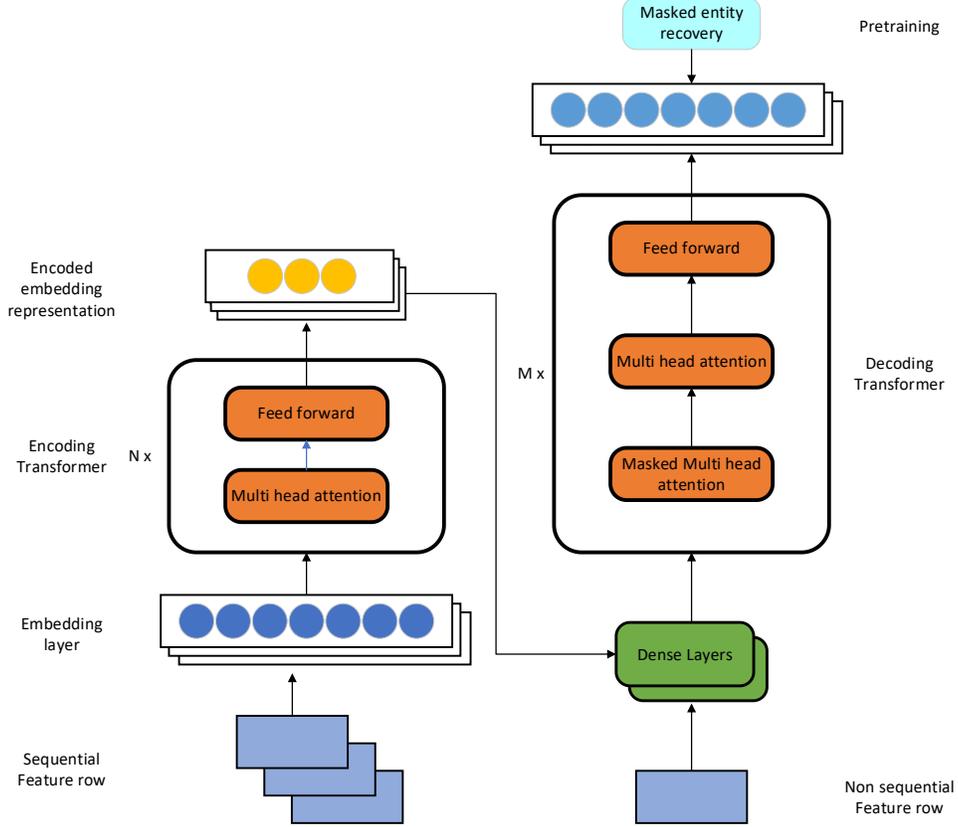}
    \caption{CASPR Transformer block}
    \label{fig:caspr}
\end{figure}
We use the transformer block popularly used for NLP tasks (\cite{vaswani2017attention}) and re-purpose it to accept activity attribute inputs as described in Figure \ref{fig:caspr}. The input to the transformer is a batch of vectors corresponding to each entity $(A_E'^1,A_E'^2...,A_E'^t)$ where each $A_E'^i = concat(i, A_{E,numerical}^i, A_{E,categorical}^i)$; $A_{E,numerical}^i$ is a vector concatenation of all numerical attributes (normalized to a standard normal distribution) of the activity at timestamp $i$ of entity $E$, and $A_{E,categorical}^i$ is a dense embedding representation obtained after passing categorical activity data through an embedding layer described in section 5.1.
The transformer encoder block uses a multi-head attention mechanism with scaled dot product attention as defined by \citet{vaswani2017attention}.
\begin{align}
    MultiHead(\mathbf{h}) = concat(head_1,...,head_n)W^O \\
    \text{where } head_i = Attention(\mathbf{h}W_i^Q,\mathbf{h}W_i^K,\mathbf{h}W_i^V) \\
    Attention(Q,K,V) = softmax(\dfrac{QK^T}{\sqrt{d_k}})V
\end{align}
where $\mathbf{h}$ is the hidden state output from the previous transformer layer. In addition to the attention mechanism layers, the output of each head is passed through a standard feed-forward neural network with 2 layers before being used as input for the next layer. \newpage Finally, we concatenate the encoded output with the non-sequential vectors $A_{E,non-sequential}$ and pass the combined output through a set of dense layers to create the final encoder output. $A_{E,non-sequential}$ vectors include customer profile information such as age, status, tenure etc.
The decoder block is almost identical to the encoder block, but also contains an additional multi-head attention layer which takes the latent representation output of the encoder as its input. We call this layer a masked layer because it masks out \textit{positions} greater than $i$, where $i$ is the sequence number of the input. This prevents the layer from including outputs that are unknown when operating at position $i$. This setup ensures the autoregressive property of the decoder while training, and can be expected to help generate better embeddings for tasks such as product recommendations, which can be thought as a similar task to predicting the next word in a sentence.
To pre-train the model on the raw input dataset, we optimize using the masked entity recovery task (\cite{vincent2008,bert2018}). Consider the input to the transformer model $(A_E'^1,A_E'^2...,A_E'^t)$. We mask out at random, 30\% of the input sequence with zero vectors for each batch of data. We then optimize the reconstruction loss by comparing the original sequence of vectors with the output of the decoder. We use the mean square error loss for numerical attributes and cross-entropy loss for the categorical attributes to measure reconstruction loss.
\subsection{Business Applications} \label{section:applications}
%
\textbf{Customer Churn} - Retaining customers is one of the most important criteria for a profitable business; the cost of acquiring new customers is generally considered 10x higher than retaining existing ones (\cite{hadden2007}). Customer churn can be defined as customers stopping doing business with a company in a given period. This could include recurrent subscribers of a product offering, or customers who make non-cyclic purchases over time. The objective of trying to predict customer churn is to allow businesses to identify at-risk customers and develop business or marketing strategies to help prevent churn. The monetary spend to retain customers is usually significantly lower than marketing spend to acquire new customers; and by using targeted marketing towards high-risk customers, businesses can maximize their return on marketing investment (\cite{mozer2000}).
For the design of our experiments, we define customers as churned when they stop interactions with the business over a defined period. An example of this definition in a retail scenario would be customers who make 0 purchases in the past 6 months. In a subscription setting, it can be defined as customers who do not renew their subscription over a period of 6 months.
%
%
\\\\
\textbf{Customer Lifetime Value} - CLV is an estimate of the expected revenue a customer generates over their association with the business. Understanding CLV has been a very useful measure for businesses to identify investment and marketing strategies, personalized retention strategies etc. (\cite{chen2018customer}). The objective of a predictive CLV model would be to accurately estimate the overall monetary value a customer would bring to the business till the point they effectively churn.
%
%
%
%
\\\\
\textbf{Product Recommendation} - Recommendation systems are useful for a variety of businesses ranging from internet ad providers, e-commerce providers, retail stores or other service providers. Understanding the best products or services desired by, or useful to specific customers helps drive business and maintain customer relationships (\cite{prodrec2007}).
Product recommendation systems primarily rely on data about users and products or services offered. User data includes historical purchases by users, interactions with products on websites or apps and profile information about the user. Product data includes metadata about the products which may include information such as brand, category, pricing etc. Popularly used methods of created recommendation systems such as collaborative filtering and latent factor methods via matrix factorization provide recommendations by grouping users and products together based on historical trends and similarities in purchasing patterns. Modern predictive methods using deep learning extends the ideas of matrix factorization with the introduction of user and product embeddings generated by deep learning models (\cite{naumovfb2019}). 
%
%
%
\\\\
\textbf{Account Fraud Detection} - Unauthorized access to user accounts or the creation of fraudulent accounts adds significant costs to the operations of a business. Some examples of added costs include abuse of new-customer discounts, rewards and other offers due to fake accounts. Additionally, there is a significant risk of stolen entitlements and resources due to unauthorized access or compromised accounts. Additionally, there is a significant risk to business reputation amongst customers due to an erosion of trust in security as well as increased friction in using and accessing account functions. The objective of an account fraud detection model is to detect fraudulent account creation and login to prevent both unauthorized access to existing accounts as well as the creation of fraud accounts.
We get a variety of customer and account metadata describing account creation and logins such as device fingerprints and merchant/business metadata. Older models used engineered statistical features based on device fingerprints, location and IP information to train a fraud detection model (\cite{roxas2011financial}). 
%
%
\section{Experiments}
The following section describes the experiments and results obtained, evaluating our framework's scalability and generalizability across real-world business tasks. We compare our framework against a traditional approach based on 
gradient-boosted decision trees, on diverse datasets, public or proprietary. The main questions we seek to answer are: (1) does CASPR representation offer a significant boost in predictive performance for any common business task? (2) how well does the CASPR embedding representation generalize across different tasks? (3) what is the most cost-effective approach to scaling-out CASPR training in an industrial setting?
\subsection{Experimental Setup}
The following describes the datasets, methods and metrics used to evaluate the generalizability of the CASPR framework in real-world business scenarios. Finally, we propose and evaluate four distinct environments to scale and distribute the main computational effort of the framework. 
\subsubsection{Datasets}
In this section, we will describe details about the datasets we used to benchmark the model performance of CASPR. A summary of the datasets can be found in Table \ref{table:datasets}.
\begin{itemize}
    \item \textbf{KKBox} is a software company which offers subscription-based music streaming services. They launched a Kaggle challenge in which the participants were asked to predict whether users will churn 30 days after their subscriptions expire. The tables they provided include transactions of users, daily user logs describing user listening behaviours, and user demographics.
    \item \textbf{Google Online Stores} is an open-source dataset from a Kaggle competition called Google Analytics Customer Revenue Prediction where the task is to predict how much each online Google Store customer will spend. The data contain user transactions from 2016 to 2018 as well as web session information such as device, geography and page visits extracted from cookies.
    \item \textbf{Instacart} data come from the ``Instacart Market Basket Analysis'' Kaggle competition where the goal is to predict which products will be in a user's next order. The dataset contains over 3 million grocery orders from more than 200,000 Instacart users.
    \item \textbf{Microsoft Retail Stores} is a Microsoft internal dataset which contains invoices, customer information, and product information for the Microsoft Retail Store.
    \item \textbf{Microsoft Accounts} is a Microsoft internal dataset which contains customer account information for the Microsoft Online Store.
    \item \textbf{Online fantasy sports platform} dataset comes from a Microsoft partner, which runs multiple popular fantasy sports leagues and betting platforms. About 4\% of all accounts in this dataset were labelled as fraud.
\end{itemize}
%
%
%
\begin{table}[h]
\centering
\caption{Summary of datasets used in experimentation (rounded for privacy)}
\begin{tabular}{lcc} 
\toprule
\textbf{Dataset} & \textbf{Size} (\# unique customers) &  \textbf{Source} \\ 
\midrule[\heavyrulewidth]
KKBox   & $\sim$1 million & public \\
\midrule
Instacart & $\sim$100k & public \\
\midrule
Google Online stores & $\sim$1 million & public \\
\midrule
Microsoft Retail Stores & $\sim$10 million & proprietary \\
\midrule
Microsoft Accounts & $\sim$1 million & proprietary \\
\midrule
Online fantasy sports platform & $\sim$1 million & proprietary  \\
\bottomrule
\end{tabular}
\label{table:datasets}
\end{table}
%
%
%
%
\subsubsection{Methods} \label{section:methods}
To generate CASPR embeddings, we used a Transformer encoder-decoder architecture with a hidden size of 16, a position-wise feed-forward dimension of 32, 6 layers, 8 self-attention heads, and a dropout probability of 0.1. We used Adam optimizer with an initial learning rate of 1e-3 to train our models. The maximum length of customer activity sequences is set to 15 and we truncated those that were longer by taking only the latest 15 transactions.
\begin{table}[h]
\centering
\caption{Summary of recency, frequency and monetary features used in baseline}



\begin{adjustbox}{max width=\textwidth}
\begin{tabular}{l|l|lll}
\toprule
\multicolumn{1}{c|}{\textbf{Recency}}    & \multicolumn{1}{c|}{\textbf{Frequency}}                                                                                 & \multicolumn{1}{c}{\textbf{Monetary}}                                                                 &  &  \\ \toprule
Time since latest activity               & Statistics of time b/w activities                                                                                       & min, max, avg, stdev                                                                                  &  &  \\ \cline{1-3}
Time since first activity                & \multirow{2}{*}{\begin{tabular}[c]{@{}l@{}}Statistics of days, weeks and \\ months when activity occurred\end{tabular}} & \multirow{2}{*}{\begin{tabular}[c]{@{}l@{}}Spending statistics \\ per week, month, year\end{tabular}} &  &  \\ \cline{1-1}
The time between first and last activity &                                                                                                                         &                                                                                                       &  &  \\ \bottomrule
\end{tabular}
\end{adjustbox}
\label{table:rfm}
\end{table}
For the baseline for the churn and customer lifetime value tasks, we used a random forest model with 100 trees and a variety of RFM features including the period since the last purchase, a number of purchases made within the period, the money spent during the period, and other aggregated statistics (e.g., min, max, standard deviation) derived from the purchase history (see Table \ref{table:rfm}) (\cite{rahim2021}). Previous work (\cite{Shwartz-Ziv2021, Chamberlain2017}) has shown that these tree ensemble models perform well across various datasets without much tuning and can be served as a strong baseline to compare with.  To evaluate the impact of CASPR on the tasks, we trained a random forest that has the same hyper-parameters as the baseline but uses CASPR embeddings as input features. We also conducted an experiment where we concatenated CASPR embeddings with RFM features and trained a random forest with the same hyperparameters as the baseline.
The baseline for detecting fraudulent accounts utilized engineered features capturing aggregated statistics on recency and frequency about different device fingerprints used, number of transactions conducted in each login session, location metadata and statistics on user impressions and clicks for each login session. These features were then used to train a LightGBM model (\cite{ke2017lightgbm}) to build the baseline. We appended CASPR embeddings of user accounts to these statistical features and trained a new LightGBM model using the same hyperparameters to identify the impact of using CASPR.
To build the baseline model for the product recommendation task, we use a collaborative filtering model using the alternating least squares (ALS) algorithm (\cite{collabf2008netflix, collabf2019spark}).
We evaluate the impact of using CASPR by generating embeddings for customers and products and using the dot product of these embeddings as an input to a similar ALS method for collaborative filtering.
%
\subsubsection{Evaluation Setup}
In this study, multiple combinations of distributed frameworks and hardware architectures are evaluated (see Table \ref{table:frameworks}). We provision and configure these environments, to identify the most cost-effective way to build CASPR models at scale. Noteworthy differences between the environments include: (1) a network gap between GPU nodes (Petastorm + Horovod environments) or (2) the NVIDIA architecture including the presence of Tensor Cores (Volta vs Kepler). The Spark-based environment includes a driver GPU which remains idle during distributed training by the design of Spark 3. With the Horovod (HVD) V100 environment, we navigate around this limitation by executing Horovod training on the GPU nodes directly. Here we still pre-load training data dynamically from a distributed storage with Petastorm. Finally, the Distributed Data-Parallel (DDP) environments aim to eliminate the network communication overhead at runtime altogether and rely on the PyTorch Distributed Data-Parallel framework, to partition the training process across GPU nodes on the same multi-GPU machine. Here the training data is loaded dynamically from local storage, practically eliminating the network communication overhead.
\begin{table}[h]
\caption{Distributed training environments evaluated for CASPR}
\centering
\begin{adjustbox}{max width=0.8\textwidth}
\begin{tabular}{l|cccc|ccc} 
\toprule
            & \multicolumn{4}{c|}{\begin{tabular}[c]{@{}c@{}}Training Distribution Frameworks\end{tabular}} & \multicolumn{3}{c}{CUDA Devices (Nodes)}  \\ 
\toprule
Environment & Spark & Horovod & Petastorm & PyTorch                                                                     & Architecture & Total & Idle               \\ 
\midrule
Spark V100  & 3.1.3   & 0.24  & 0.11.4 & 1.10.2                                                                      & V100 (Volta)        & 5     & 1                  \\ 
\midrule
HVD V100    & -     & 0.24  & 0.11.4  & 1.10.2                                                                      & V100 (Volta)        & 4     & 0                  \\ 
\midrule
DDP K80     & -     & -    & -    & 1.10.2                                                                       & K80 (Kepler)      & 4     & 0                  \\ 
\midrule
\textbf{DDP V100}    & -     & -    & -    & \textbf{1.10.2}                                                                       & \textbf{V100 (Volta)}        & \textbf{4}     & \textbf{0}                  \\
\bottomrule
\end{tabular}
\end{adjustbox}
\label{table:frameworks}
\end{table}
\subsection{Predictive Performance} \label{section:prediction}
%
We evaluated a diverse set of tasks ranging from predicting customer churn, estimating lifetime value, detecting fraudulent accounts and ranking product recommendations. For all the classification tasks, we use the Area under the Receiver Operating Characteristics curve (AUROC) as our primary comparison metric. For the task of predicting customer churn, we additionally report the F1-score of the churning class. We also report the root-mean-squared error (RMSE) for the task of estimating customer lifetime value. For the ranking task of product recommendation, we report the Mean Average Precision (MAP), Precision@1, Success@5 and Normalized Discounted Cumulative Gain@3 (NDCG@3). We note that in comparison with the baseline model, CASPR shows improvement across most tasks and metrics over a variety of datasets; public, first party and Microsoft partners.
In the experiments for predicting customer churn on the KKBox and Microsoft Retail Store datasets, we find that CASPR improves the AUROC score by 2-3 points, with larger improvements in the F1 score (see Table \ref{table:classificationchurn}). Interestingly, CASPR does not show tangible improvements in the Google Online Stores dataset. One of the key factors for these results is the sparsity of the Google Online Stores dataset; customers have an average of 1.5 past activities. Because of the very short activity history, CASPR does not learn a lot of patterns of sequences, unlike the KKBox dataset which has an average activity history of 15.
\begin{table}[h]
\caption{CASPR representation impact on predicting customer churn}
\centering
\begin{adjustbox}{max width=0.9\textwidth}
\begin{tabular}{lcccccc} 
\cmidrule[\heavyrulewidth]{1-5}
\multicolumn{1}{l|}{Classification Task} & \multicolumn{1}{c|}{Dataset} & \multicolumn{1}{c|}{Representation} & \multicolumn{1}{c|}{AUROC} & \multicolumn{1}{c}{F1}   &  &   \\ 
\cmidrule[\heavyrulewidth]{1-5}
\multicolumn{1}{c|}{\multirow{7}{*}{Customer Churn}}    & \multicolumn{1}{c|}{\multirow{2}{*}{KKBox}}   & \multicolumn{1}{c}{Baseline}           & \multicolumn{1}{|c|}{0.89} & \multicolumn{1}{c}{0.27} &  &   \\ 
\cline{3-5}
\multicolumn{1}{l|}{}                    & \multicolumn{1}{c|}{}        & \multicolumn{1}{c}{CASPR}       & \multicolumn{1}{|c|}{\textbf{0.91}} & \multicolumn{1}{c}{\textbf{0.44}} &  &   \\ 

\cline{2-5}
\multicolumn{1}{c|}{}      & \multicolumn{1}{c|}{\multirow{2}{*}{Google Online Stores}}   & \multicolumn{1}{c}{Baseline}           & \multicolumn{1}{|c|}{0.897} & \multicolumn{1}{c}{0.96} &  &   \\ 
\cline{3-5}
\multicolumn{1}{l|}{}                    & \multicolumn{1}{c|}{}        & \multicolumn{1}{c}{CASPR}       & \multicolumn{1}{|c|}{0.903} & \multicolumn{1}{c}{0.96} &  &   \\

\cline{2-5}
\multicolumn{1}{c|}{}      & \multicolumn{1}{c|}{\multirow{3}{*}{Microsoft Retail Stores}}   & \multicolumn{1}{c}{Baseline}           & \multicolumn{1}{|c|}{0.761} & \multicolumn{1}{c}{0.814} &  &   \\ 
\cline{3-5}
\multicolumn{1}{l|}{}                    & \multicolumn{1}{c|}{}        & \multicolumn{1}{c}{CASPR}       & \multicolumn{1}{|c|}{0.777} & \multicolumn{1}{c}{0.831} &  &   \\
\cline{3-5}
\multicolumn{1}{l|}{}                    & \multicolumn{1}{c|}{}        & \multicolumn{1}{c}{CASPR w/RFM features}       & \multicolumn{1}{|c|}{\textbf{0.794}} & \multicolumn{1}{c}{\textbf{0.837}} &  &   \\

\cmidrule[\heavyrulewidth]{1-5}
\end{tabular}
\label{table:classificationchurn}
\end{adjustbox}
\end{table}
For the estimation of the customer's lifetime value with the Microsoft Retail Stores dataset, using CASPR gets an improvement of 2.5 points in the AUROC score, which is similar to the improvement seen when using CASPR for predicting churn on the Microsoft Retail Stores dataset (see Table \ref{table:classificationclv}).
\begin{table}[h]
\caption{CASPR impact on predicting customer lifetime value. We have segmented the data into high- and low-value customers using the Pareto Principle (top customers that generated 80\% of the revenue were considered as high-value) and calculated the AUROC score using these segment labels}
\centering
\begin{adjustbox}{max width=0.9\textwidth}
\begin{tabular}{lcccccc} 
\cmidrule[\heavyrulewidth]{1-5}
\multicolumn{1}{l|}{Regression Task} & \multicolumn{1}{c|}{Dataset} & \multicolumn{1}{c|}{Representation} & \multicolumn{1}{c|}{AUROC} & \multicolumn{1}{c}{RMSE}   &  &   \\ 
\cmidrule[\heavyrulewidth]{1-5}
\multicolumn{1}{c|}{\multirow{2}{*}{Customer Lifetime Value}}      & \multicolumn{1}{c|}{\multirow{2}{*}{Microsoft Retail Stores}}   & \multicolumn{1}{c}{Baseline}           & \multicolumn{1}{|c|}{0.659} & \multicolumn{1}{c}{1108} &  &   \\ 
\cline{3-5}
\multicolumn{1}{l|}{}                    & \multicolumn{1}{c|}{}        & \multicolumn{1}{c}{CASPR}       & \multicolumn{1}{|c|}{\textbf{0.685}} & \multicolumn{1}{c}{\textbf{1103}} &  &   \\
\cmidrule[\heavyrulewidth]{1-5}
\end{tabular}
\label{table:classificationclv}
\end{adjustbox}
\end{table}
On a very different task of detecting fraudulent accounts, we also see significant improvements over the baseline model with the use of CASPR embeddings. These results validate our claim of the generalizability of using CASPR embeddings across a variety of tasks (see Table \ref{table:classification}).
\begin{table}[h]
\caption{CASPR representation impact on detecting fraudulent accounts}
\centering
\begin{adjustbox}{max width=0.9\textwidth}
\begin{tabular}{lccccc} 
\cmidrule[\heavyrulewidth]{1-4}
\multicolumn{1}{l|}{Classification Task} & \multicolumn{1}{c|}{Dataset} & \multicolumn{1}{c|}{Representation} & \multicolumn{1}{c}{AUROC}  &  \\ 
\cmidrule[\heavyrulewidth]{1-4}
\multicolumn{1}{c|}{\multirow{4}{*}{Digital Account Fraud Detection}}      & \multicolumn{1}{c|}{\multirow{2}{*}{Online fantasy sports platform}}   & \multicolumn{1}{c}{Baseline}           & \multicolumn{1}{|c}{0.811}  &    \\
\cline{3-4}
\multicolumn{1}{l|}{}                    & \multicolumn{1}{c|}{}        & \multicolumn{1}{c}{CASPR}       & \multicolumn{1}{|c}{\textbf{0.883}} &   \\
\cline{2-4}
\multicolumn{1}{c|}{}      & \multicolumn{1}{c|}{\multirow{2}{*}{Microsoft Accounts Data}}   & \multicolumn{1}{c}{Baseline}           & \multicolumn{1}{|c}{0.873}  &    \\
\cline{3-4}
\multicolumn{1}{l|}{}                    & \multicolumn{1}{c|}{}        & \multicolumn{1}{c}{CASPR}       & \multicolumn{1}{|c}{\textbf{0.895}}  &   \\
\cmidrule[\heavyrulewidth]{1-4}
\end{tabular}
\label{table:classification}
\end{adjustbox}
\end{table}
In the ranking task of product recommendation using CASPR shows a significant improvement over the collaborative filtering baseline on the Instacart dataset (see Table \ref{table:prodrec}). We believe this relative improvement in model metrics can be partly explained by the density of the Instacart data; each customer has a long activity history with an average of over 20 purchases. The existence of relatively longer event sequences works well with the transformer design of the CASPR model.
\begin{table}[h]
\caption{CASPR representation impact on product recommendation ranking performance}
\centering
\begin{adjustbox}{max width=0.9\textwidth}
\begin{tabular}{l|c|c|c|c|c|c} 
\cmidrule[\heavyrulewidth]{1-7}
Ranking Task           & Dataset & Representation & MAP  & Prec @1 & Success @5 & NDCG @3  \\ 
\cmidrule[\heavyrulewidth]{1-7}
\multirow{4}{*}{Product Recommendation} & \multirow{2}{*}{Instacart}   & Baseline           & 0.21  & 0.32    & 0.61    & 0.28     \\ 
\cline{3-7}
                       &         & CASPR       & \textbf{0.46} & \textbf{0.62}    & \textbf{1.46}    & \textbf{0.56}     \\
\cline{2-7}
 & \multirow{2}{*}{Microsoft Retail Stores}   & Baseline           & 0.13  & 0.09    & 0.24    & 0.11     \\ 
\cline{3-7}
                       &         & CASPR       & \textbf{0.14} & \textbf{0.10}    & \textbf{0.27}    & \textbf{0.12}       \\
\cmidrule[\heavyrulewidth]{1-7}
\end{tabular}
\label{table:prodrec}
\end{adjustbox}
\end{table}
\subsection{Training at Scale} \label{section:scale}
\begin{figure}[h]
    \centering
    \includegraphics[scale=0.8]{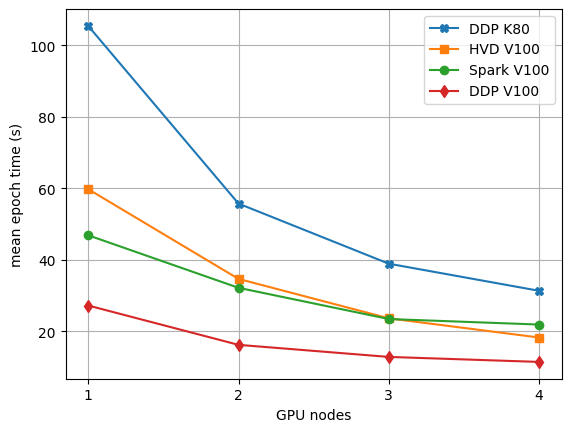}
    \caption{CASPR pre-training cost depending on the choice and size of the training environment. Number of epochs = 10, dataset = KKBox, batch size = 8192}
    \label{fig:scalability}
\end{figure}

%
We conduct a series of scalability experiments by executing the same CASPR model training (i.e., using the same hyperparameters and the same dataset) across different  computation environments described in Table \ref{table:frameworks}. 
Every experiment consists of 10-epoch training on the KKBox dataset. A batch size of 8192 was assumed to prevent running out of CUDA memory in the environment with the smallest GPU(s). Figure \ref{fig:scalability} shows the average epoch duration depending on the choice and size of the distribution environment. %
\begin{table}[h]
\caption{CASPR training duration, depending on the choice and the size of the environment for distributed training. Number of epochs = 10, dataset = KKBox, batch size = 8192. We measure average epoch time, overall training duration and the total GPU time, as proxies of the training cost.}
\centering
\begin{adjustbox}{max width=0.8\textwidth}
\begin{tabular}{l|r|r|r|r} 
\cmidrule[\heavyrulewidth]{1-5}
\textbf{Environment}                & \multicolumn{1}{l|}{\textbf{GPUs}} & \multicolumn{1}{l|}{\textbf{training time (s)}} & \multicolumn{1}{l|}{\textbf{epoch time (s)}} & \multicolumn{1}{c}{\textbf{GPU time (s)}}  \\ 
\cmidrule[\heavyrulewidth]{1-5}
\multirow{4}{*}{Spark V100}    & 1                                  & 468.62                                        & 46.86                     & 937.24                                       \\ 
\cline{2-5}
                                    & 2                                  & 321.50                                           & 32.15                                                          & 964.50                                          \\ 
\cline{2-5}
                                    & 3                                  & 234.12                                        & 23.41                                                         & 936.48                                       \\ 
\cline{2-5}
                                    & 4                                  & 218.64                                        & 21.86                                                               & 1093.21                                      \\ 
\cmidrule[\heavyrulewidth]{1-5}
\multirow{4}{*}{HVD V100} & 1                                  & 597.65                                         & 59.76                    & 597.65                                        \\ 
\cline{2-5}
                                    & 2                                  & 346.36                                         & 34.63                                                          & 692.73                                         \\ 
\cline{2-5}
                                    & 3                                  & 236.03                                         & 23.60                                                         & 708.11                                        \\ 
\cline{2-5}
                                    & 4                                  & 182.31                                         & 18.23                                                        & 729.22                                        \\ 
\cmidrule[\heavyrulewidth]{1-5}
\multirow{4}{*}{\textbf{DDP V100}}       & 1                                  & 271.91                                        & 27.19                      & 271.91                                       \\ 
\cline{2-5}
                                    & 2                                  & 161.78                                        & 16.17                                                         & 323.56                                       \\ 
\cline{2-5}
                                    & 3                                  & 128.01                                          & 12.80                                                      & 384.03                                         \\ 
\cline{2-5}
                                    & 4                                  & 114.02                                         & 11.40                                                       & 456.10                                        \\ 
\cmidrule[\heavyrulewidth]{1-5}
\multirow{4}{*}{DDP K80}        & 1                                  & 1054.71                                       & 105.47                    & 1054.71                                      \\ 
\cline{2-5}
                                    & 2                                  & 557.28                                        & 55.73                                                        & 1114.57                                      \\ 
\cline{2-5}
                                    & 3                                  & 388.99                                        & 38.90                                                         & 1166.98                                      \\ 
\cline{2-5}
                                    & 4                                  & 313.07                                        & 31.31                                                         & 1252.27                                      \\ 
\cmidrule[\heavyrulewidth]{1-5}
\end{tabular}
\end{adjustbox}
\label{table:distribution}
\end{table}
It is noteworthy, that the distributed environment managed by Spark and Horovod, proves least efficient among the evaluated environments. At the time of writing this paper, Spark framework still requires the driver node to be equipped with an identical GPU device as the worker node, however utilizes only the worker nodes for distributed training w/ Horovod (leaving the driver node idle). GPU training can only be as fast as the data feeding effort at each iteration. The additional network layer (or even geographical distance) between GPU worker nodes, can limit the throughput further. The remaining environments configured for this study seek to eliminate the scheduling and communication inefficiencies wherever possible. The Horovod (HVD) V100 environment still includes a network gap between the GPU nodes, however all of the nodes are now utilized. Training performance is comparable with the Spark V100 environment, however the cost decreases by 20\% (see: GPU(s) time, Table \ref{table:distribution}). Finally the last two Distributed Data-Parallel (DDP) environments distribute the training workload across GPU nodes available within the same machine, eliminating the network gap and limiting the communication overhead. We observe a 4x cost reduction after switching from Spark to DDP. We also observe a speed up of between 3x and 4x after switching from NVIDIA Kepler to Volta architecture, in the same experiments. Application of PyTorch Distributed Data-Parallel framework on a multi-GPU machine with a recent NVIDIA architecture emerges as the most efficient approach to scaling CASPR training in an industrial setting.
\section{Conclusions}
In this study, we have proposed CASPR, a novel framework for computing deep embedding representation of customer activity data represented in tabular format. We have conducted a diverse set of experiments to evaluate its generalization potential across common business optimization tasks and scalability to real-world enterprise workloads. The predictive experiments show a significant boost in performance, over traditional feature representation and suggest generalizability across different business tasks. The scale-out experiments offer an evaluation of operationalization alternatives for large-scale industrial production. We evaluate popular frameworks for distributed deep learning, and identify the most cost effective way of learning CASPR representation from tabular big data.
%
%
\section{Acknowledgments}
This project is supported by the Business Applications \& Platform team within Cloud + AI division, Microsoft Corporation. We would like to thank Pushpraj Shukla and Walter Sun.
\clearpage
\bibliographystyle{plainnat}
{\small\bibliography{bibliography}}

\begin{thebibliography}{34}
\providecommand{\natexlab}[1]{#1}
\providecommand{\url}[1]{\texttt{#1}}
\expandafter\ifx\csname urlstyle\endcsname\relax
  \providecommand{\doi}[1]{doi: #1}\else
  \providecommand{\doi}{doi: \begingroup \urlstyle{rm}\Url}\fi

\bibitem[Ahmad et~al.(2019)Ahmad, Jafar, and Aljoumaa]{Ahmad2019}
Abdelrahim~Kasem Ahmad, Assef Jafar, and Kadan Aljoumaa.
\newblock Customer churn prediction in telecom using machine learning in big
  data platform.
\newblock \emph{Journal of Big Data}, 6, 12 2019.
\newblock ISSN 21961115.
\newblock \doi{10.1186/s40537-019-0191-6}.

\bibitem[Aljunid and Manjaiah(2019)]{collabf2019spark}
Mohammed~Fadhel Aljunid and D.~H. Manjaiah.
\newblock Movie recommender system based on collaborative filtering using
  apache spark.
\newblock In Valentina~Emilia Balas, Neha Sharma, and Amlan Chakrabarti,
  editors, \emph{Data Management, Analytics and Innovation}, pages 283--295,
  Singapore, 2019. Springer Singapore.
\newblock ISBN 978-981-13-1274-8.

\bibitem[Arik and Pfister(2019)]{Arik2019}
Sercan~O. Arik and Tomas Pfister.
\newblock Tabnet: Attentive interpretable tabular learning.
\newblock 8 2019.

\bibitem[Barkan and Koenigstein(2016)]{Barkan2016}
Oren Barkan and Noam Koenigstein.
\newblock Item2vec: Neural item embedding for collaborative filtering.
\newblock 3 2016.

\bibitem[Borisov et~al.(2021)Borisov, Leemann, Seßler, Haug, Pawelczyk, and
  Kasneci]{Borisov2021}
Vadim Borisov, Tobias Leemann, Kathrin Seßler, Johannes Haug, Martin
  Pawelczyk, and Gjergji Kasneci.
\newblock Deep neural networks and tabular data: A survey.
\newblock 10 2021.

\bibitem[Chamberlain et~al.(2017)Chamberlain, Ângelo Cardoso, Liu, Pagliari,
  and Deisenroth]{Chamberlain2017}
Benjamin~Paul Chamberlain, Ângelo Cardoso, C.~H.~Bryan Liu, Roberto Pagliari,
  and Marc~Peter Deisenroth.
\newblock Customer lifetime value prediction using embeddings.
\newblock volume Part F129685, pages 1753--1762. Association for Computing
  Machinery, 8 2017.
\newblock ISBN 9781450348874.
\newblock \doi{10.1145/3097983.3098123}.

\bibitem[Chen et~al.(2018)Chen, Guitart, del R{\'\i}o, and
  Peri{\'a}nez]{chen2018customer}
Pei~Pei Chen, Anna Guitart, Ana~Fern{\'a}ndez del R{\'\i}o, and Africa
  Peri{\'a}nez.
\newblock Customer lifetime value in video games using deep learning and
  parametric models.
\newblock In \emph{2018 IEEE international conference on big data (big data)},
  pages 2134--2140. IEEE, 2018.

\bibitem[Chui et~al.(2018)Chui, Manyika, Miremadi, Henke, Chung, Nel, and
  Malhotra]{Chui2018}
Michael Chui, James Manyika, Mehdi Miremadi, Nicolaus Henke, Rita Chung, Pieter
  Nel, and Sankalp Malhotra.
\newblock Notes from the ai frontier insights from hundreds of use cases.
\newblock \emph{McKinsey Global Institute}, 2018.

\bibitem[Devlin et~al.(2018)Devlin, Chang, Lee, and Toutanova]{bert2018}
Jacob Devlin, Ming{-}Wei Chang, Kenton Lee, and Kristina Toutanova.
\newblock {BERT:} pre-training of deep bidirectional transformers for language
  understanding.
\newblock \emph{CoRR}, abs/1810.04805, 2018.
\newblock URL \url{http://arxiv.org/abs/1810.04805}.

\bibitem[Fader et~al.(2005)Fader, Hardie, and Lee]{Fader2005}
Peter~S. Fader, Bruce~G.S. Hardie, and Ka~Lok Lee.
\newblock Rfm and clv: Using iso-value curves for customer base analysis.
\newblock \emph{Journal of Marketing Research}, 42:\penalty0 415--430, 11 2005.
\newblock ISSN 0022-2437.
\newblock \doi{10.1509/jmkr.2005.42.4.415}.

\bibitem[Grbovic et~al.(2016)Grbovic, Radosavljevic, Djuric, Bhamidipati,
  Savla, Bhagwan, and Sharp]{Grbovic2016}
Mihajlo Grbovic, Vladan Radosavljevic, Nemanja Djuric, Narayan Bhamidipati,
  Jaikit Savla, Varun Bhagwan, and Doug Sharp.
\newblock E-commerce in your inbox: Product recommendations at scale.
\newblock 6 2016.
\newblock \doi{10.1145/2783258.2788627.}

\bibitem[Gruener et~al.(2018)Gruener, Cheng, and Litvin]{Gruener2018}
Robbie Gruener, Owen Cheng, and Yevgeni Litvin.
\newblock Introducing petastorm: Uber atg’s data access library for deep
  learning.
\newblock \emph{Uber Engineering Blog}, 2018.

\bibitem[Hadden et~al.(2007)Hadden, Tiwari, Roy, and Ruta]{hadden2007}
John Hadden, Ashutosh Tiwari, Rajkumar Roy, and Dymitr Ruta.
\newblock Computer assisted customer churn management: State-of-the-art and
  future trends.
\newblock \emph{Computers \& Operations Research}, 34\penalty0 (10):\penalty0
  2902--2917, 2007.
\newblock ISSN 0305-0548.
\newblock \doi{https://doi.org/10.1016/j.cor.2005.11.007}.
\newblock URL
  \url{https://www.sciencedirect.com/science/article/pii/S0305054805003503}.

\bibitem[Huang et~al.(2020)Huang, Khetan, Cvitkovic, and Karnin]{Huang2020}
Xin Huang, Ashish Khetan, Milan Cvitkovic, and Zohar Karnin.
\newblock Tabtransformer: Tabular data modeling using contextual embeddings.
\newblock 12 2020.

\bibitem[Ke et~al.(2017)Ke, Meng, Finley, Wang, Chen, Ma, Ye, and
  Liu]{ke2017lightgbm}
Guolin Ke, Qi~Meng, Thomas Finley, Taifeng Wang, Wei Chen, Weidong Ma, Qiwei
  Ye, and Tie-Yan Liu.
\newblock Lightgbm: A highly efficient gradient boosting decision tree.
\newblock \emph{Advances in neural information processing systems},
  30:\penalty0 3146--3154, 2017.

\bibitem[Kowalczyk and Larsen(2018)]{kowalczyk2018scalable}
Damian~Konrad Kowalczyk and Jan Larsen.
\newblock Scalable privacy-compliant virality prediction on twitter.
\newblock \emph{arXiv preprint arXiv:1812.06034}, 2018.

\bibitem[Le and Mikolov(2014)]{Le2014}
Quoc~V. Le and Tomas Mikolov.
\newblock Distributed representations of sentences and documents.
\newblock 5 2014.

\bibitem[Li et~al.(2020)Li, Zhao, Varma, Salpekar, Noordhuis, Li, Paszke,
  Smith, Vaughan, Damania, and Chintala]{Li2020}
Shen Li, Yanli Zhao, Rohan Varma, Omkar Salpekar, Pieter Noordhuis, Teng Li,
  Adam Paszke, Jeff Smith, Brian Vaughan, Pritam Damania, and Soumith Chintala.
\newblock Pytorch distributed: Experiences on accelerating data parallel
  training.
\newblock 6 2020.

\bibitem[Mozer et~al.(2000)Mozer, Wolniewicz, Grimes, Johnson, and
  Kaushansky]{mozer2000}
M.C. Mozer, R.~Wolniewicz, D.B. Grimes, E.~Johnson, and H.~Kaushansky.
\newblock Predicting subscriber dissatisfaction and improving retention in the
  wireless telecommunications industry.
\newblock \emph{IEEE Transactions on Neural Networks}, 11\penalty0
  (3):\penalty0 690--696, May 2000.
\newblock ISSN 1941-0093.
\newblock \doi{10.1109/72.846740}.

\bibitem[Naumov et~al.(2019)Naumov, Mudigere, Shi, Huang, Sundaraman, Park,
  Wang, Gupta, Wu, Azzolini, Dzhulgakov, Mallevich, Cherniavskii, Lu,
  Krishnamoorthi, Yu, Kondratenko, Pereira, Chen, Chen, Rao, Jia, Xiong, and
  Smelyanskiy]{naumovfb2019}
Maxim Naumov, Dheevatsa Mudigere, Hao{-}Jun~Michael Shi, Jianyu Huang,
  Narayanan Sundaraman, Jongsoo Park, Xiaodong Wang, Udit Gupta, Carole{-}Jean
  Wu, Alisson~G. Azzolini, Dmytro Dzhulgakov, Andrey Mallevich, Ilia
  Cherniavskii, Yinghai Lu, Raghuraman Krishnamoorthi, Ansha Yu, Volodymyr
  Kondratenko, Stephanie Pereira, Xianjie Chen, Wenlin Chen, Vijay Rao, Bill
  Jia, Liang Xiong, and Misha Smelyanskiy.
\newblock Deep learning recommendation model for personalization and
  recommendation systems.
\newblock \emph{CoRR}, abs/1906.00091, 2019.
\newblock URL \url{http://arxiv.org/abs/1906.00091}.

\bibitem[Rahim et~al.(2021)Rahim, Mushafiq, Khan, and Arain]{rahim2021}
Mussadiq~Abdul Rahim, Muhammad Mushafiq, Salabat Khan, and Zulfiqar~Ali Arain.
\newblock Rfm-based repurchase behavior for customer classification and
  segmentation.
\newblock \emph{Journal of Retailing and Consumer Services}, 61:\penalty0
  102566, 2021.
\newblock ISSN 0969-6989.
\newblock \doi{https://doi.org/10.1016/j.jretconser.2021.102566}.
\newblock URL
  \url{https://www.sciencedirect.com/science/article/pii/S0969698921001326}.

\bibitem[Roxas(2011)]{roxas2011financial}
Maria~L Roxas.
\newblock Financial statement fraud detection using ratio and digital analysis.
\newblock \emph{Journal of Leadership, Accountability, and Ethics}, 8\penalty0
  (4):\penalty0 56--66, 2011.

\bibitem[Santoro et~al.(2018)Santoro, Fiano, Bertoldi, and
  Ciampi]{santoro2018big}
Gabriele Santoro, Fabio Fiano, Bernardo Bertoldi, and Francesco Ciampi.
\newblock Big data for business management in the retail industry.
\newblock \emph{Management Decision}, 2018.

\bibitem[Schmittlein et~al.(1987)Schmittlein, Morrison, and
  Colombo]{Schmittlein1987}
David~C. Schmittlein, Donald~G. Morrison, and Richard Colombo.
\newblock Counting your customers: Who-are they and what will they do next?
\newblock \emph{Management Science}, 33:\penalty0 1--24, 1 1987.
\newblock ISSN 0025-1909.
\newblock \doi{10.1287/mnsc.33.1.1}.

\bibitem[Sergeev and Balso(2018)]{Sergeev2018}
Alexander Sergeev and Mike~Del Balso.
\newblock Horovod: fast and easy distributed deep learning in tensorflow.
\newblock 2 2018.

\bibitem[Shwartz-Ziv and Armon(2021)]{Shwartz-Ziv2021}
Ravid Shwartz-Ziv and Amitai Armon.
\newblock Tabular data: Deep learning is not all you need.
\newblock 6 2021.

\bibitem[Somepalli et~al.(2021)Somepalli, Goldblum, Schwarzschild, Bruss, and
  Goldstein]{Somepalli2021}
Gowthami Somepalli, Micah Goldblum, Avi Schwarzschild, C.~Bayan Bruss, and Tom
  Goldstein.
\newblock Saint: Improved neural networks for tabular data via row attention
  and contrastive pre-training.
\newblock 6 2021.

\bibitem[Vanderveld et~al.(2016)Vanderveld, Pandey, Han, and
  Parekh]{Vanderveld2016}
Ali Vanderveld, Addhyan Pandey, Angela Han, and Rajesh Parekh.
\newblock An engagement-based customer lifetime value system for e-commerce.
\newblock volume 13-17-August-2016, pages 293--302. Association for Computing
  Machinery, 8 2016.
\newblock ISBN 9781450342322.
\newblock \doi{10.1145/2939672.2939693}.

\bibitem[Vaswani et~al.(2017)Vaswani, Shazeer, Parmar, Uszkoreit, Jones, Gomez,
  Kaiser, and Polosukhin]{vaswani2017attention}
Ashish Vaswani, Noam Shazeer, Niki Parmar, Jakob Uszkoreit, Llion Jones,
  Aidan~N. Gomez, Lukasz Kaiser, and Illia Polosukhin.
\newblock Attention is all you need.
\newblock \emph{CoRR}, abs/1706.03762, 2017.
\newblock URL \url{http://arxiv.org/abs/1706.03762}.

\bibitem[Vincent et~al.(2008)Vincent, Larochelle, Bengio, and
  Manzagol]{vincent2008}
Pascal Vincent, Hugo Larochelle, Yoshua Bengio, and Pierre-Antoine Manzagol.
\newblock Extracting and composing robust features with denoising autoencoders.
\newblock In \emph{Proceedings of the 25th International Conference on Machine
  Learning}, ICML '08, page 1096–1103, New York, NY, USA, 2008. Association
  for Computing Machinery.
\newblock ISBN 9781605582054.
\newblock \doi{10.1145/1390156.1390294}.
\newblock URL \url{https://doi.org/10.1145/1390156.1390294}.

\bibitem[Xiao and Benbasat(2007)]{prodrec2007}
Bo~Xiao and Izak Benbasat.
\newblock E-commerce product recommendation agents: Use, characteristics, and
  impact.
\newblock \emph{MIS Quarterly}, 31\penalty0 (1):\penalty0 137--209, 2007.
\newblock ISSN 02767783.
\newblock URL \url{http://www.jstor.org/stable/25148784}.

\bibitem[Zaharia et~al.(2012)Zaharia, Chowdhury, Das, Dave, Ma, McCauly,
  Franklin, Shenker, and Stoica]{Zaharia2012}
Matei Zaharia, Mosharaf Chowdhury, Tathagata Das, Ankur Dave, Justin Ma, Murphy
  McCauly, Michael~J. Franklin, Scott Shenker, and Ion Stoica.
\newblock Resilient distributed datasets: A {Fault-Tolerant} abstraction for
  {In-Memory} cluster computing.
\newblock In \emph{9th USENIX Symposium on Networked Systems Design and
  Implementation (NSDI 12)}, pages 15--28, San Jose, CA, April 2012. USENIX
  Association.
\newblock ISBN 978-931971-92-8.
\newblock URL
  \url{https://www.usenix.org/conference/nsdi12/technical-sessions/presentation/zaharia}.

\bibitem[Zaharia et~al.(2016)Zaharia, Xin, Wendell, Das, Armbrust, Dave, Meng,
  Rosen, Venkataraman, Franklin, Ghodsi, Gonzalez, Shenker, and
  Stoica]{Zaharia2016}
Matei Zaharia, Reynold~S. Xin, Patrick Wendell, Tathagata Das, Michael
  Armbrust, Ankur Dave, Xiangrui Meng, Josh Rosen, Shivaram Venkataraman,
  Michael~J. Franklin, Ali Ghodsi, Joseph Gonzalez, Scott Shenker, and Ion
  Stoica.
\newblock Apache spark.
\newblock \emph{Communications of the ACM}, 59:\penalty0 56--65, 10 2016.
\newblock ISSN 0001-0782.
\newblock \doi{10.1145/2934664}.

\bibitem[Zhou et~al.(2008)Zhou, Wilkinson, Schreiber, and
  Pan]{collabf2008netflix}
Yunhong Zhou, Dennis Wilkinson, Robert Schreiber, and Rong Pan.
\newblock Large-scale parallel collaborative filtering for the netflix prize.
\newblock In Rudolf Fleischer and Jinhui Xu, editors, \emph{Algorithmic Aspects
  in Information and Management}, pages 337--348, Berlin, Heidelberg, 2008.
  Springer Berlin Heidelberg.
\newblock ISBN 978-3-540-68880-8.

\end{thebibliography}
\end{document}